\def\year{2020}\relax
\documentclass[letterpaper]{article} 
\usepackage{aaai20}  
\usepackage{times}  
\usepackage{helvet} 
\usepackage{courier}  
\usepackage[hyphens]{url}  
\usepackage{graphicx} 
\usepackage{graphicx}  
\usepackage{mathtools}
\usepackage{makecell}
\usepackage{amssymb}
\usepackage{amsmath}
\usepackage{kotex}
\usepackage{multirow,hhline,array}
\usepackage[usenames, dvipnames]{xcolor}
\newcommand{\nj}[1]{\textcolor{black}{#1}}
\newcommand{\sw}[1]{\textcolor{black}{#1}}
\title{FEED: Feature-level Ensemble for Knowledge Distillation}

\author{
SeongUk Park, Nojun Kwak\\
Graduate School of Convergence Science and Technology\\
  Seoul National University\\
  \texttt{\{swpark0703, nojunk\} @snu.ac.kr} \\
}
 
\begin{document}

\maketitle

\begin{abstract}
Knowledge Distillation (KD) aims to transfer knowledge in a teacher-student framework, by providing the predictions of the teacher network to the student network in the training stage to help the student network generalize better. It can use either a teacher with high capacity or {an} ensemble of multiple teachers.
However, the latter is not convenient when one wants to use feature-map-based distillation methods.
For a solution, this paper proposes a versatile and powerful training algorithm named FEature-level Ensemble for knowledge Distillation (FEED), which aims to transfer the ensemble knowledge using multiple teacher networks.
We introduce a couple of training algorithms that transfer ensemble knowledge to the student at the feature map level.
{Among the feature-map-based distillation methods, using several non-linear transformations in parallel for transferring the knowledge of the multiple teacher{s} helps the student find more generalized solutions.}
We name this method as parallel FEED, and
experimental results on CIFAR-100 and ImageNet show that our method has clear performance enhancements, without introducing any additional parameters or computations at test time. 
We also show the experimental results of {sequentially feeding teacher's information to the student, hence the name sequential FEED,} and discuss the lessons obtained.
Additionally, the empirical results on measuring the reconstruction errors at the feature map give hints for the enhancements.
\end{abstract}

\noindent \section{Introduction}
Recent successes of convolutional neural networks (CNNs) have led to the use of deep learning in real-world applications.
In order to manipulate these deep learning models, deep CNNs are trained using multi-class datasets to find manifolds separating different classes well.
To meet this need, deep and parameter-rich networks have emerged that have the power to find manifolds for {a large number} of classes.
However, these deep CNNs suffer from the problem of overfitting due to their great depth and complexity, which results in a drop of performance at the test time.
In fact, even a small ResNet applied for a dataset such as CIFAR-100 \cite{cifar100} will easily overfit 
with the converged train losses, whereas the test accuracy is significantly {lower}.
These phenomena have led to the need for learning DNN models with appropriate regularization to allow them to generalize better. 
Regularizing a model to achieve high performance for new inputs is a technique that has been used since the era of early machine learning.

\begin{figure}[t]

    \includegraphics[width=0.46\textwidth]{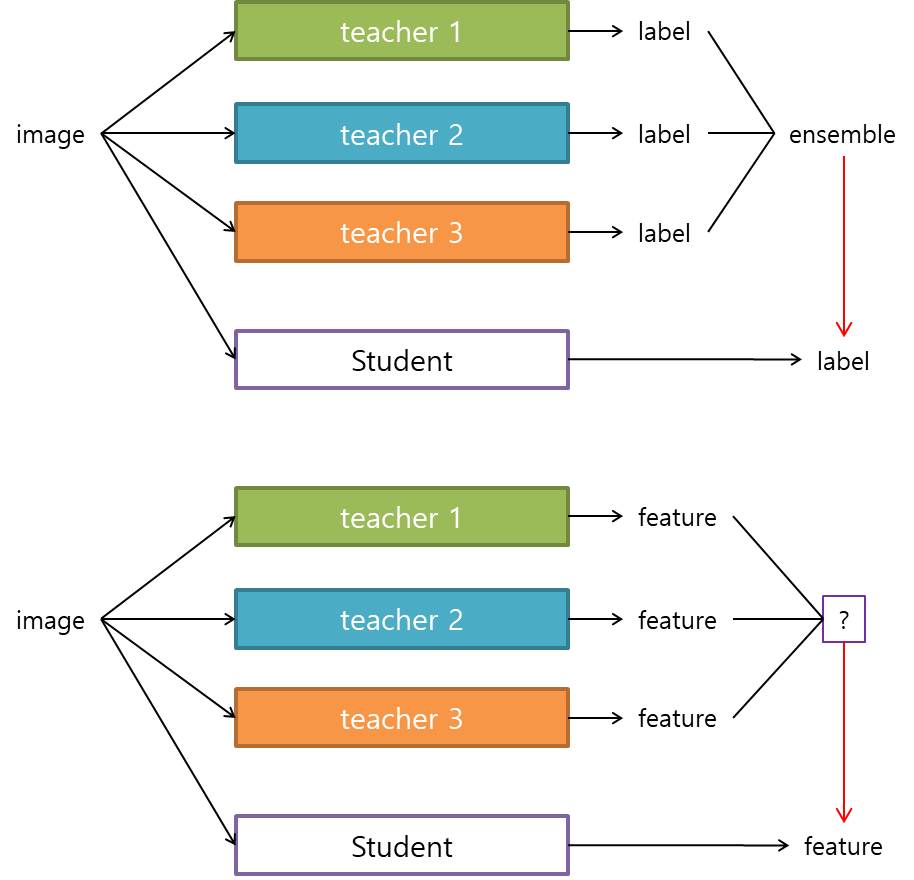}
    \caption{Our problem formulation. The figure at the top shows how the KD is trained using ensemble of networks as the teacher. 
    The figure at the bottom shows that a problem arises when we want to distill ensemble information directly in feature level rather than using label information. 
    \label{fig:introduction}}

\end{figure}

Model ensemble \cite{dietterich2000ensemble} is one of the popular regularization methods, which  has been used as a way of alleviating the problem of overfitting in a single model. 
However, it has drawbacks in that it requires multiple models and inputs should be fed to each of them at the test time.
Many studies proposed ideas to transfer knowledge of {a} teacher to {a} compact student~\cite{ba2014deep,rusu2015policy,modelcompression,urban2016deep}.
For a solution to this problem, \cite{hinton2015distilling} proposed {\textit{Knowledge Distillation}} (KD) which trains a student network using soft labels {from} an ensemble of multiple models or a {teacher network with high capacity}. 
{They obtained} meaningful results in the speech recognition problem, and KD has become one of the representative methods of {\textit{knowledge transfer}}.
{They {aim} performance improvements of a weak student network by giving various forms of knowledge of expert teacher networks.}
It is also categorized as one family of model compression since it helps the student network achieve higher accuracy with a fixed number of parameters. 

The recent knowledge transfer {algorithms} can be approximately categorized in two {viewpoints}. 
The first is whether to use an ensemble model or a single high-capacity model as a teacher. 
{The second is whether the teacher delivers predicted labels or the information from feature maps.}
Whereas the methods that use teacher's prediction can use both types of teachers, to the best of our knowledge, methods using {feature-map-level} information can only use a single high-capacity model as a teacher.
For example, studies of {\textit{Factor Transfer} (FT) \cite{kim2018paraphrasing}, \textit{Attention Transfer}  (AT)  \cite{zagoruyko2016paying}, and \textit{Neuron Selectivity Transfer}}  (NST) \cite{huang2017like} set the student network as {a} {shallow} {one} with a small number of parameters, and set the teacher network as {a} deeper and more powerful {one} instead of {an} ensemble expert.
One of the drawbacks of methods with a high-capacity teacher model is that high-capacity model may be hard to obtain \cite{lan2018knowledge}.

On the contrary, the ones which use {an} ensemble of networks can transfer ensemble knowledge and also have advantages of the peer-teaching framework \cite{furlanello2018born,hinton2015distilling,lan2018knowledge}.
Also, \cite{zhang2017deep} showed that using {the} same type of network {for transferring output-level knowledge} can improve the performance of a network.
{However, the methods that deliver knowledge at the feature-map level also have advantages that they can give more specific information to the student compared to the methods that only rely on the output predictions of {the} teacher.} 

{To make full advantages of both the ensemble teacher method and the feature map transferring} method, we propose a new framework that {delivers} knowledge of multiple {networks} at {the} feature-map-level (See Fig. \ref{fig:introduction}). 
In this paper, we train a new student network using multiple teachers that share the same architecture with the student network, named parallel FEED.
We also explore a {variant}, named sequential FEED, which recursively trains a new teacher. 
Comparisons with other algorithms are provided with analysis that provides lessons. 
Our main {contributions} are threefold:
{
\begin{itemize}
\item In various settings, we empirically show that for high-capacity networks, feature-map-based methods that give more specific knowledge are stronger than label-based methods which give more abstract knowledge.
\item We propose parallel FEED, a method that allows multiple teacher networks to be used for knowledge distillation at the feature-map level with non-linear transformations.
\item For qualitative analysis, {by utilizing the autoencoder reconstruction losses, we provide} hints for the {performance} enhancements {of our FEED.}
\end{itemize}
}

{The paper is organized as follows.} 
First, we briefly explain the related works in the area of knowledge distillation. 
Then our main proposed method is described 
and a variant version implemented to compare with other methods is proposed. 
Next, we verify our proposed methods with experiments. 
{Experimental} results from our proposed training methods are compared with AT, KD, FT, BAN on CIFAR-100. 
The ImageNet dataset is also used to check the feasibility of our method on a large-scale dataset. 
Qualitative analysis is also provided which is followed by discussion and conclusion.

\section{Related Works}
Many researchers studied the ways to train models other than {using a} purely supervised loss. 
In the early times of these studies, \textit{Model Compression} \cite{modelcompression} studied the ways to compress information from ensemble models in one network. 
Ba \cite{ba2014deep} showed that shallow feed-forward nets can learn the complex functions previously learned by deep neural nets, by {minimizing} ${L}_{2}$ loss between logits.  

More recently, Hinton \cite{hinton2015distilling} proposed \textit{Knowledge Distillation} {(KD)}, which uses softened softmax labels from teacher networks in training the student network {by minimizing the following loss}:
\begin{equation}
\small
\mathcal{L}_{KD}=(1-\alpha)\mathcal{L}_{CE}(y, \sigma(\textbf{s})) + \alpha T^{2}\mathcal{L}_{KL}\bigg(\sigma\big(\frac{\textbf{s}}{T}\big), \sigma\big(\frac{\textbf{t}}{T}\big)\bigg),
\label{eq:L_KD-likeFT}
\end{equation}
where the $\sigma(\cdot)$ is the softmax function, $T$ is a temperature value that controls the softened logit. $\alpha$ is hyper-parameter that controls the weight between two terms. 
The vectors $\textbf{s}$ and $\textbf{t}$ are predicted output logits of the student network and the teacher network, respectively, and $y$ is the ground-truth label.
$\mathcal{L}_{CE}$ is the cross-entropy loss that is commonly used in classification problems, and $\mathcal{L}_{KL}$ is the Kullback-Leibler divergence loss.

This motivated researchers to develop many variants of it to various domains, and many researchers studied ways to better teach the student \cite{frosst2017distilling,pham2018efficient,radosavovic2018data,tan2018learning,tarvainen2017mean}.
We introduce some recent research flows that are potentially related to our proposed method.

\noindent \textbf{Peer teaching framework:} 
\sw{Rather recently, many papers adapt peer-teaching framework that use the same kind of network for \nj{both the} teacher and \nj{the} student.} Geras \cite{geras2015blending} try to transfer knowledge between two networks that have almost the same number of parameters. 
BAN \cite{furlanello2018born} shows that using the exact same architecture for the teacher and the student boosts the performance of the student network even without softening the labels. 
%
{BAN uses a simpler loss term without softening both logits which KD does, and does not even assign weights to the two terms as follows:}
\begin{equation}
\mathcal{L}_{BAN}=
\mathcal{L}_{CE}\big(y, \sigma(\textbf{s})\big) + 
\mathcal{L}_{KL}\big(\sigma(\textbf{t}), \sigma({\textbf{s}})\big). 
\label{eq:BAN}
\end{equation}
%
They train the student recursively to enhance the performance further. The $n^{th}$ student network becomes the $(n+1)^{th}$ teacher network to train the next student network. The better teacher network will {teach the student network better}.

Also, studies such as DML \cite{zhang2017deep} and ONE \cite{lan2018knowledge} use the same kind of network {for on-line training of peer networks} with mutual KL-divergence losses. 

\noindent \textbf{Feature-map-based methods for knowledge transfer:}
Contrary to the methods that try to use labels from the teacher network, there exist studies that distill useful information directly from feature maps in various forms. 

AT \cite{zagoruyko2016paying} tried to transfer the attention map of the teacher network to {the} student network, and got meaningful results in knowledge transfer and transfer learning tasks. Their loss term is:
\begin{equation}
\small
\mathcal{L}_{AT}=\mathcal{L}_{CE}\big(y, \sigma(\textbf{s})\big) + \beta\sum^{L}_{l=1}\|\frac{f({A}^{t}_{l})}{\|f({A}^{t}_{l})\|_2}-\frac{f({A}^{s}_{l})}{\|f({A}^{s}_{l})\|_2} \|_2 ,
\label{eq:L_KD-likeFT}
\end{equation}
where $\beta$ is a hyperparameter that depends on the number of elements, and $l$ denotes the $l^{th}$ group \cite{yosinski2014transferable} in the network. $A^{t}_{l}$, $A^{s}_{l}$ are attention maps obtained from the teacher network and the student network, and $f(A) = (1/N)\sum^{N}_{n=1}a^{2}_{n}$, where $N$ is the number of channels, and $a_n$ is the spatial map from the $n^{th}$ channel.

Yim \cite{yim2017gift} {introduced another} knowledge transfer technique for faster optimization and applied it also to transfer learning. Shen\cite{shen2019amalgamating} tried to combine knowledge trained from different domains, and You \cite{You:2017:LMT:3097983.3098135} utilized ensemble of orderings of samples to teach the student network, which is very novel.
Huang \cite{huang2017like} tried to match the {features of the student and teacher networks by} 
devising a loss term MMD (Maximum Mean Discrepancy). 

FT \cite{kim2018paraphrasing} uses additional paraphraser and translator networks which help training the student network and got meaningful results. 
Their loss terms are:
\begin{equation}
	\mathcal{L}_{rec} = \| x-P(x) \|^2_2, \quad {\text{(for the paraphraser)}} 
    \label{eq:L_AE}
\end{equation} 
\begin{equation}
\mathcal{L}_{student}=\mathcal{L}_{CE}(y, \sigma(\textbf{s})) + \beta\|\frac{F_T}{\|F_T\|_2}-\frac{F_S}{\|F_S\|_2} \|_1,
\label{eq:L_student}
\end{equation}
where $P(\cdot)$ is the autoencoder-based paraphraser network and $x$ is the input feature map for the paraphraser. 
$F_T$ and $F_S$ are the output of the paraphraser and the translator, respectively.

\section{Proposed Training {Algorithms}}
In deep CNNs, due mainly to the curse of dimensionality, the data points that lie on the data space are very sparse. {For example, CIFAR datasets that contains the number of 50,000 training images and has 3,072 dimensions, so distances between each samples are very far.}
{Necessarily, decision boundaries that determine the borders dividing {classes} are multitudinous, because finding boundaries that {fit} well to {a training} dataset is {relatively} an easy task. 
{Even if the {networks with the same architecture} are trained, the learned decision boundaries cannot be the same.}}
This is why ensemble methods usually {perform} better than a single model despite their structural equality. 
{Goodfellow} \cite{Goodfellow-et-al-2016} also state that different models will not make all the same errors on the test dataset.

Consider the conditions that {determine} the training procedure of {CNNs}. 
{They include} the structure of the CNN and the choice {of an} optimizer, the seed of random initialization, the sequence of mini-batches, and the {types of data} augmentations.
If one makes the same conditions for two different CNNs, their training procedure will be identical. 
However, we usually determine only the structure of the CNN, usually keeping others to be random. 
Consequently, two networks with {the} same structure will definitely not learn the same decision boundaries.

{Additionally,} {Kim \cite{kim2018paraphrasing}} {state} that they resolve the `inherent difference' between two networks. 
Among the inherent differences, minimizing the differences in the structure of CNNs can help better learn the knowledge of the teacher network.
This has a thread of connection with that of {BAN \cite{furlanello2018born}}, which {shows} that using the same architecture for both {the} student and {the} teacher is actually beneficial.
This motivation provides us chances to produce several modified versions of existing methods.
In this section, we explain the {feature-level ensemble} training algorithms that are used for boosting the performance of {a} student network without introducing any {additional} calculations at {the} test time. 
The proposed method is named as FEED which is an abbreviation for the \textit{FEature-level Ensemble for knowledge Distillation}.
We propose pFEED (\textit{parallel FEED}), which we handle as our main method, that use non-linear transformation layers to distill ensemble knowledge into the student network. 
We also {introduce} sFEED (\textit{sequential FEED}), motivated by BAN \cite{furlanello2018born}, which adapts the sequential training with the use of nonlinear transformations. 
The use of the nonlinear layers, rather than {using} a simple distance metric, had been explored previously in FitNet \cite{romero2014fitnets} and FT \cite{kim2018paraphrasing}.

\begin{figure}[t]
\centering
\includegraphics[width=1\linewidth]{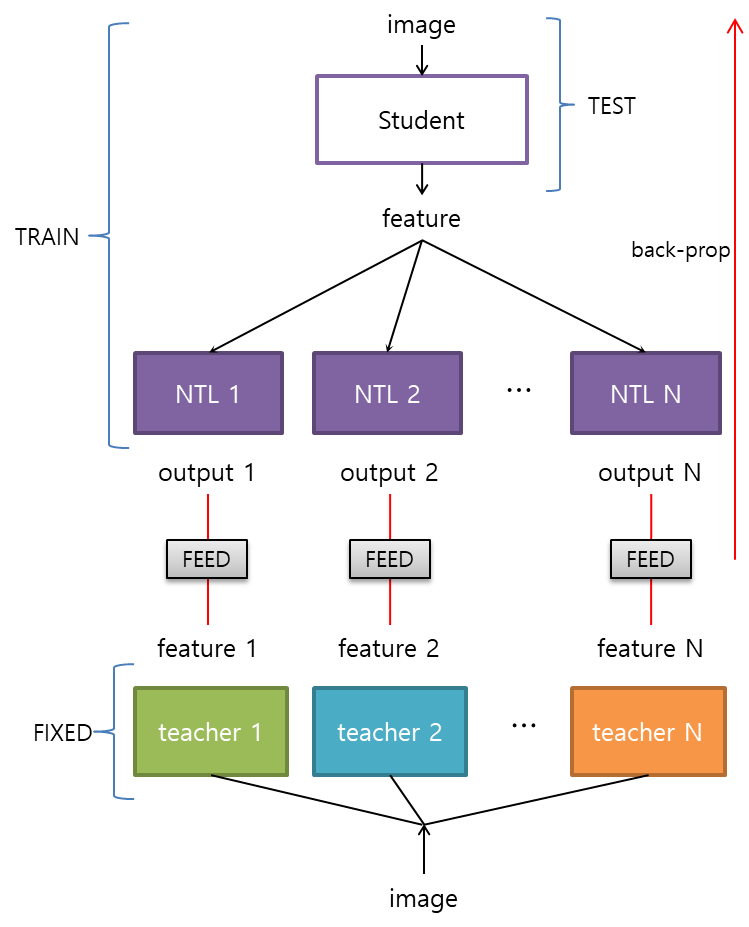}\\
\caption{Illustration of our proposed {method}, parallel FEED. {NTL is an} abbreviation of Nonlinear Transformation Layer, and {one NTL is} allocated to each teacher {network}. All teacher networks are fixed during the training, and the student network and NTL networks are {trained simultaneously}.}
\label{pfeed}
\end{figure}

\subsection{{Parallel FEED}}\label{sec:pfeed}

Assuming that {distilling} knowledge {at the} feature map level and {distilling} knowledge {from an ensemble of multiple networks} both have their advantages, we wanted to make cooperation of these two kinds of methods.
Tackling this problem, we propose a training structure named \textit{pFEED} to transfer the ensemble knowledge at the feature map level.
Let us denote the number of the teacher network as ${N}$.
We use {different} non-linear transformation layers {for} each {teacher}, which means if there are ${N}$ teachers, 
there are ${N}$ different nonlinear layers.
The final feature map of the student network is fed into the non-linear layers and its output is trained to mimic the final feature map of the teacher network.
In this way, we take {advantages} of both the ensemble model and the feature-based method.
Our {method} is illustrated in Figure \ref{pfeed}.
If we use ${N}$ different teachers, the loss term {is as follows:}
\begin{equation}
\mathcal{L}_{student}=\mathcal{L}_{CE}(y, \sigma(\textbf{s})) + \beta\sum^{N}_{n=1}{\mathcal{L}_{FEED_{n}}},
\label{eq:L_KD-likeFT}
\end{equation}
\begin{equation}
\mathcal{L}_{FEED_{n}} = \|{\frac{x^T_{n}}{\|x^T_{n}\|_2}-\frac{{NTL_n}(x^S)}{\|{NTL_n}(x^S)\|_2}}\|_1.
\label{eq:L_FT}
\end{equation} 
{Here, $\mathcal{L}_{CE}$ is the} cross-entropy loss, $y$ {is the} ground-truth label, $\textbf{s}$ {is} the predicted logits, {and $\sigma(\cdot)$ denotes the softmax function.} $\mathcal{L}_{FEED_{n}}$ is the FEED loss from {the} $n^{th}$ teacher network, $x^T_n$ is the output feature map obtained from {the} $n^{th}$ teacher network, and ${x^S}$ is the output feature map obtained from the student network. ${NTL_n}(\cdot)$ is {the} $n^{th}$ nonlinear transformation layer used for {adapting the student with the} $n^{th}$ teacher network. \sw{Each ${NTL_n}(\cdot)$ is composed of three convolution layers with \nj{the} kernel size of 3 to expand the size of receptive field, so that the student can flexibly merge the knowledge attained from different teachers.}
The feature maps are normalized by its own size as in (\ref{eq:L_FT}). This normalizing term was previously used in AT. {$\beta$} is used to scale the $\mathcal{L}_{1}$ distance loss to match the scale of $\mathcal{L}_{CE}$, {as also described in AT}.

\begin{figure}[t]
\centering
\includegraphics[width=1\linewidth]{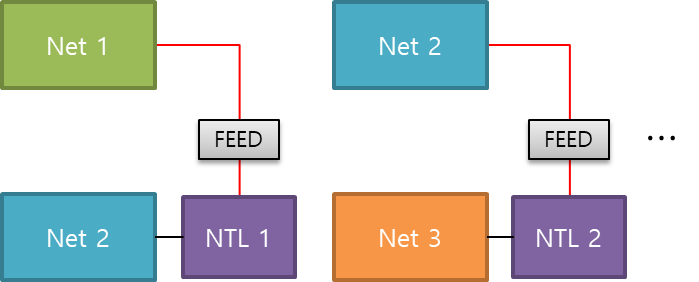}\\
\caption{The sequential FEED. We adapted the setting of BANs that the teacher network and the student network have the same architecture.
 The trained student network is used as a teacher network for the next {stage}. 
{Dots mean} that we repeat this procedure.
}
\label{fig:sfeed}
\end{figure}

\subsection{{Sequential FEED}}\label{sec:sfeed}
BAN \cite{furlanello2018born} used cross-entropy loss combined with KD loss without softening the softmax logits.
They use a trained student network as a new teacher {which is used to train a new student} and do this recursively.
{We take this architectural advantage of using the same type of network recursively} 
because it {is a suitable model for accumulating and assembling knowledge.} 
{By} performing knowledge transfer several times recursively, it may also ensemble knowledge of many training sequences.
We applied the FEED recursively and named this framework as \textit{sFEED (sequential FEED)}. 
{The training procedure of sFEED} is illustrated in Figure \ref{fig:sfeed}. 

If the student network is trained standalone, it would perform similarly to the teacher network. 
However, from the view of knowledge ensemble, since {the} teacher network delivers {feature-level} knowledge different from {that of the} student network, the student network will benefit from it.

\section{Experiments}
\label{sec:exp}
Firstly, we want to briefly show with experiments that using non-linear transformations for the output feature map with a distance metric is helpful for teaching the student network, rather than just {using} a simple distance metric {without any adaptation layer}. 
Next, we will report the score of pFEED, which use multiple pairs of nonlinear transformation layers with each teacher networks, and will compare ours with KD and AT in {a} similar setting.
Finally, sFEED will be compared with FT, KD, and BAN. 
The algorithms will be experimented following the BAN's sequential training schedules.

We show the classification results on CIFAR-100 \cite{cifar100} {on} which many networks show lower test accuracy than train accuracy that many studies do experiments on it to show their regularization power. 
On this dataset, our results are compared with feature map based methods and label based methods with {corresponding} settings. 
Second, we explore the feasibility of our algorithm on Imagenet \cite{ILSVRC15}, a commonly used large dataset and analyze the results quantitatively. In the remaining section, we show some analysis of our algorithm. The implementation details are on the supplementary material.

We chose three types of CNNs to check the applicability of our algorithms on CIFAR-100: ResNet \cite{he2016deep}, Wide ResNet \cite{zagoruyko2016wide}, and ResNext \cite{xie2017aggregated}. 
For ResNets, we chose ResNet-56 and ResNet-110 which have fewer number of parameters compared to recent CNNs. WRN28-10 is a model that controls the widen factor, with much more number of parameters. 
WRN28-10 achieves the best classification accuracy on CIFAR-100 among the WRNs reported {in \cite{zagoruyko2016wide}}. 
The ResNext29-16x64d also achieves the best classification accuracy on CIFAR-100 in {\cite{xie2017aggregated}. 
This} type of CNNs controls the cardinality of CNNs, and it has much more parameters compared to other models.
For ImageNet, we used ResNet-34 to confirm the feasibility {of our method} on large scale {datasets}.

\subsection{{Effectiveness of FEED loss}}
We compare the classification results using our FEED loss and two other {kinds} of loss terms, AT and simple $\mathcal{L}_{1}$ loss, {in training} the student at {the feature map level}.
The results are shown {in} Table \ref{TOY}.
The $\mathcal{L}_{1}$ means we simply use $\mathcal{L}_{1}$ loss at the final feature maps.
AT use attention maps attained from feature maps to give information.
From these {results}, we can see that using nonlinear transformation layers are helpful {in} delivering information at the feature-map level.
Interestingly, AT and $\mathcal{L}_{1}$ {beats} {a} single FEED model in ResNext29-16x64d model.

\begin{table}[t]
\caption{
Test classification error on {Cifar-100} dataset. 
The numbers on the \textbf{scratch} column {are the} baseline errors of each network, trained by the pure cross-entropy loss, which {are the scores} of the teachers. The numbers on \textbf{scratch*} columns are the reported errors on their original papers.
}
\resizebox{1.\linewidth}{!}{
{\renewcommand{\arraystretch}{1.2}
\begin{tabular}{l|cc|ccc}
\Xhline{3\arrayrulewidth}
\multicolumn{1}{c|}{Model Type} & scratch*& scratch & $\mathcal{L}_{1}$  & AT    & FEED  \\
\hline 
ResNet-56                       & -& 28.18   & 27.16              & 26.60 & 26.02 \\
ResNet-110                      & -& 26.97   & 25.42              & 25.70 & 25.25 \\
WRN28-10                        & 19.25& 19.09   & 17.94              & 17.86 & 17.68 \\
ResNext29-16x64d                & 17.31& 17.32   & 16.46                  & 16.51 & 16.80 \\
\Xhline{3\arrayrulewidth}
\end{tabular}
}
}
\label{TOY}
\end{table}


\subsection{Parallel FEED}\label{ex:parallel}
In {this experiment} on {pFEED}, our main experiment, we used the same type of networks that were used on previous experiments. 
For all four types of CNNs, we compared the classification results with the result of KD because we designed our training algorithm with the intention of {distilling} more ensemble-like knowledge from multiple teachers. 
{We also experimented AT to empirically show that it would not be easy for a feature-map-based knowledge transfer methods to fully utilize multiple teacher networks}. 
We modulated AT to use multiple teachers, by simply incrementally adding same loss terms for each teacher network. We did not change the $\beta$ values for weights of each loss terms, {similar} to pFEED.

The results are in Table \ref{pfeed}. 
The `Scratch' column shows the performance of the base networks, {which are} used in KD for model ensemble, and also used as teachers in pFEED and AT. 
For all experiments, pFEED consistently got higher accuracy compared with both KD and AT and produces the closest {results} to {those of the} network ensemble. 

\begin{table*}[]
\caption{{Test classification error (\%) on CIFAR-100 dataset. All scores of other methods are our reproduction. In the 5x$\mathcal{L}_{1}$, 5xAT and 5xFT columns, we used five $\mathcal{L}_{1}$, AT and FT losses each for training one student.}
The {scores of the Ens} column {are the} performance of label ensemble of 5 scratch models. We trained pFEED 5 times and averaged the results.}
\begin{center}
\begin{tabular}{l|c|ccccccc}
\Xhline{3\arrayrulewidth}
\multicolumn{1}{c|}{Model Type}           &\multicolumn{1}{c|}{Scratch (5 mean)}   & KD & 5x$\mathcal{L}_{1}$    &  5xAT   & 5xFT & pFEED  &  Ens & Parameters \\ 
\hline
ResNet-56            & 28.18 ($\pm$ 0.17) & \textbf{24.69} & 27.29 & 27.02 & 25.41 & 24.74 ($\pm$ 0.12) & 22.45 & 0.85M\\
ResNet-110           & 26.97 ($\pm$ 0.16) & 23.50 & 28.94 & 25.23 & 23.81 & \textbf{22.98} ($\pm$ 0.20) & 21.20 & 1.73M\\
WRN28-10             & 19.09 ($\pm$ 0.13) & 18.30 & 23.75 & 17.73 & 17.30 & \textbf{16.86} ($\pm$ 0.14) & 16.59 & 36.5M\\
ResNext29-16x64d     & 17.32 ($\pm$ 0.08) & 16.64 & 16.31 & 17.42 & 16.30 & \textbf{15.70} ($\pm$ 0.08) & 15.66 & 68.1M\\
\Xhline{3\arrayrulewidth}
\end{tabular}
\end{center}
\label{pfeed}
\end{table*}

\textbf{Comparison with KD:} It is worth noting that the performance of KD is almost equivalent to pFEED for small networks, but as it comes to the networks with {a} larger number of parameters, pFEED shows better accuracy compared to KD. 
This result matches the hypothesis that {distilling in the feature-map level} will {provide} more \textbf{detailed} information {to the student}. 
{In contrast, {KD} works quite well for small networks, because {it gives} ensemble labels, which {are} rather \textbf{abstract}. 
These labels are useful for small networks that should focus on key information {for accuracy improvement}.}

\sw{\textbf{Comparison with feature-map-based methods:}  Compare the results on Table \ref{pfeed} with Table \ref{TOY}. As shown by the scores of $\mathcal{L}_{1}$ and AT, using multiple teachers \nj{with these methods} did not make a meaningful difference compared to the case when a single teacher network was used.}
{This is} {an example} of our statement, that it is not easy for existing {feature-map-based} knowledge transfer methods to {fully} utilize multiple teachers to boost the performance of a single student network.

The results of pFEED for ImageNet is on Table \ref{table:pfeed-imagenet}. 
We also could find some accuracy improvements on ImageNet dataset, but did not have enough resources to train models with larger parameters, nor did we could experiment on larger models or other methods. 
Though we reported scores of five scratch models, only {the} first three teacher networks had been used for training {of the student}. 
We could get decent results, but interestingly, improvements are not as strong as those of sFEED on ImageNet that will be shown later.

\begin{table}[t]
\caption{Validation classification error (\%) of pFEED on Imagenet dataset.}
\resizebox{1.\linewidth}{!}{
{\renewcommand{\arraystretch}{1.2}
\begin{tabular}{l|ccccc|c}
\Xhline{3\arrayrulewidth}
\multicolumn{1}{c|}{Model Type}  &\multicolumn{5}{c|}{Scratch (5 runs)} & pFEED \\ 
\hline
ResNet-34(Top-1)   & 26.45    & 26.59 & 26.40 & 26.77 & 26.64 & \textbf{25.27}  \\
ResNet-34(Top-5)   & 8.54     &  8.72 &  8.63 &  8.68 &  8.61 & \textbf{7.79}  \\
\Xhline{3\arrayrulewidth}
\end{tabular}
}}
\label{table:pfeed-imagenet}
\end{table}

\subsection{Sequential FEED}\label{sec:compare}
This section contains results of 4 types of {methods (FT, KD, BAN, and sFEED)} experimented in small {and large} capacity networks.
Note that the pFEED is our main proposed method for good performance, and {the} purpose of {this subsection} is to further delve into each method to speculate the characteristics and differences of each {method}.
The classification results of {different methods} on CIFAR-100 can be found in Table \ref{table:sfeed_comp}. 
The word `Stack' in {Table} \ref{table:sfeed_comp} is the number of recursions that the student model is trained.
We only experimented up to 5 times since all of them {achieve} fairly good enough accuracy compared to baseline models.
We reported the {performance of} sFEED because it has a more similar framework to BANs, though the pFEED, our main proposed method, outperforms {sFEED}.
The BAN-N in {\cite{furlanello2018born}} would be {the} identical setting to the Stack-(N-1) in Table \ref{table:sfeed_comp}.
We additionally experimented FT with {the identically structured teacher and student networks}.
FT basically uses {a} large teacher network {with} a paraphraser, which is trained as an autoencoder. 
It is trained to extract key information called `factor' from the teacher network in {an} unsupervised {manner}, so it gives more abstract information.

\begin{table*}[]
\caption{Test classification error (\%) of sFEED on CIFAR-100 dataset. The model's scores on the \textbf{Scratch*} column are the same as the scores reported on their original papers, and {those} on the \textbf{Scratch} column are from our implementation. The parameters are counted in Millions.}
\begin{center}
\resizebox{0.7\linewidth}{!}{
\begin{tabular}{c|l|c|c|c|ccccc}
\Xhline{3\arrayrulewidth}
\multirow{16}*{CIFAR-100}& \multicolumn{1}{c|}{Model Type}  & Scratch             & Distillation Type & Algorithm & Stack-2& Stack-3& Stack-4& Stack-5&  \\
\cline{2-10}
                        & \multirow{4}*{ResNet-56}          & \multirow{4}*{28.03}& \multirow{2}*{Label}   & BAN        &25.85 & 25.52 & \textbf{25.30} & 25.45 & \\
                        &                                   &                     &                        & KD         &25.55 & 25.14 & \textbf{24.98} & 24.98 & \\
\cline{4-5}
                        &                                   &                     & \multirow{2}*{Feature} & FT         &26.33 & 25.70 & 25.91 & \textbf{25.18} & \\
                        &                                   &                     &                        & sFEED      &26.02 & 26.00 & 25.59 & \textbf{25.33} & \\
\cline{2-10}
                        & \multirow{4}*{ResNet-110}         & \multirow{4}*{27.14}& \multirow{2}*{Label}   & BAN        &24.67 & \textbf{23.81} & 23.98 & 24.06 & \\
                        &                                   &                     &                        & KD         &24.35 & 24.01 & 23.73 & \textbf{23.66} & \\
\cline{4-5}
                        &                                   &                     & \multirow{2}*{Feature} & FT         &24.90 & 24.50 & 24.34 & \textbf{24.13} & \\
                        &                                   &                     &                        & sFEED      &25.25 & \textbf{24.33} & 24.58 & 24.40 & \\
\cline{2-10}
                        & \multirow{3}*{WRN28-10}           & \multirow{3}*{19.00}& \multirow{1}*{Label}   & KD         &18.47 & \textbf{18.43} & 18.57 & 19.05 & \\
\cline{4-5}
                        &                                   &                     & \multirow{2}*{Feature} & FT         &18.23 & 18.05 & 18.14 & \textbf{17.84} & \\
                        &                                   &                     &                        & sFEED      &17.68 & 17.50 & 17.52 & \textbf{17.27} & \\
\cline{2-10}
                        & \multirow{4}*{ResNext29-16x64d}   & \multirow{4}*{17.31}& \multirow{2}*{Label}   & BAN        &16.59 & \textbf{16.42} & 16.43 & 16.48 & \\
                        &                                   &                     &                        & KD         &\textbf{16.87} & 16.90 & 16.88 & - & \\
\cline{4-5}
                        &                                   &                     & \multirow{2}*{Feature} & FT         &16.80 & 16.76 & \textbf{16.47} & 16.48 & \\
                        &                                   &                     &                        & sFEED      &16.80 & 16.47 & 16.22 & \textbf{15.94} & \\
\Xhline{3\arrayrulewidth}
\end{tabular}
}
\end{center}
\label{table:sfeed_comp}
\end{table*}

{\textbf{Comparison of Label-based-methods and Feature-map-based methods:} 
For the smaller networks such as ResNet-56 and ResNet 110, methods using labels performed better than the {feature-map-based} methods, but for networks with larger sizes, {feature-map-based} methods showed higher accuracy. 
FT {uses} more abstract knowledge (using paraphraser) compared to sFEED, so it did not perform well as sFEED for larger networks. {However,} it performed better for smaller-sized networks.
KD use{s} more abstract knowledge compared to BAN, because KD soften{s} the labels, and it achieve{s} higher accuracy for smaller networks, but BAN showed better accuracy for ResNext.}

The results of {sFEED} for ImageNet is on Table \ref{ex-imagenet}. 
For the base model, we simply used the pre-trained model that Pytorch supplies, and could achieve {the desired} result that the performance of Top-1 and Top-5 accuracy improves at each {Stack}. 
The sFEED with Stack-5 achieves better performance compared to pFEED. {The performance of} pFEED in Table \ref{table:pfeed-imagenet} trained with only three teachers {is} close to {that of} sFEED with Stack-3, which is a reasonable comparison.

\begin{table}[]
\caption{Validation classification error (\%) of sFEED on Imagenet dataset. The model's scores on the \textbf{Scratch*} column are the same as the scores reported on the Pytorch implementation.}
\resizebox{1.\linewidth}{!}{
{\renewcommand{\arraystretch}{1.2}
\begin{tabular}{l|cccccc}
\Xhline{3\arrayrulewidth}
\multicolumn{1}{c|}{Model Type}  &\multicolumn{1}{c}{Scratch*} & Stack-2 & Stack-3 & Stack-4 & Stack-5 
\\ \hline
ResNet-34(Top-1)   & 26.45    & 25.60 & 25.30 & 25.18 & \textbf{25.00}  \\
ResNet-34(Top-5)   & 8.54     & 8.08  &  7.86 &  \textbf{7.73} &  7.83  \\
\Xhline{3\arrayrulewidth}
\end{tabular}
}}
\label{ex-imagenet}
\end{table}

\subsection{Qualitative Analysis}\label{recon}
\textbf{Reconstruction Loss:} 
{Suppose} that the reason for the accuracy gains shown in the previous tables is that the student learns the ensemble knowledge 
{that contains information with high complexity.}
But how can one actually distinguish whether the networks learn complex information or not?
Here, we adopt the paraphraser from FT.
A paraphraser in the FT is a convolutional autoencoder in that it uses convolution {and} transposed convolution layers {with a} reconstruction loss. Then the factor can be interpreted as a latent vector ${z}$. 

Let us denote the input of paraphraser as {$x$}. The increase in the complexity of feature representation is equivalent to the increase in the complexity of $x$. 
Since the number of parameters in {the} paraphraser is fixed, the size of $z$ also should be fixed. 
Consequently, {as the complexity of $x$ increases}, $p(x|z)$ decreases, resulting in an increase of {the} reconstruction loss. 
Reconstruction errors were used as a criterion for feature selection or PCAs in  \cite{boutsidis2008unsupervised,farahat2011efficient,li2017reconstruction,masaeli2010convex}, where they use linear models. 
In our experiment, we use an arbitrary autoencoder composed of convolution layers with nonlinear activation.
\begin{figure}[t]
\centering

    {\includegraphics[width=0.3\textwidth]{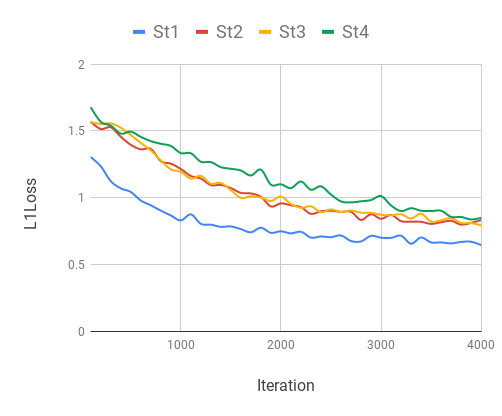}
    \caption{Paraphraser reconstruction loss $\mathcal{L}_{rec}$(training) for ResNet-56 with sFEED.}\label{rec_loss_step}}
    {\includegraphics[width=0.3\textwidth]{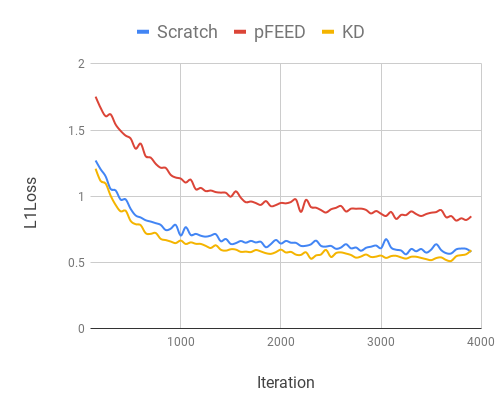}
    \caption{Paraphraser reconstruction loss $\mathcal{L}_{rec}$(training) for ResNet-56 with pFEED.}\label{rec_loss_kd}}

\end{figure}

For sFEED and pFEED, we recorded the {average training} reconstruction losses of the paraphrasers {for ResNet-56} {normalized by the size of the paraphraser} and plotted the curve on Figure \ref{rec_loss_step} and Figure \ref{rec_loss_kd} (though we do not actually use paraphrasers for FEED training). 
{In Fig. \ref{rec_loss_step}, St1} through St4 are the paraphrasers trained based on the student networks {of} Scratch {(St1)} through Stack-4 {(St4)} in Table \ref{table:sfeed_comp}. 
The paraphrasers in Figure \ref{rec_loss_kd} is trained based on one of the scratch teacher networks and following student network {of pFEED {and KD}} in Table \ref{pfeed}. 
{As expected, as the knowledge is transferred, the reconstruction loss becomes larger which indicates that the student network learns more difficult knowledge and thus the classifier accuracy increases.}
{This} trend matches the results on the tables. 
In Figure \ref{rec_loss_step}, the {the trend of reconstruction losses reciprocally matches the accuracy trend} of ResNet-56 row in Table \ref{table:sfeed_comp} of sFEED.
{(Especially, Stack 2 and 3 have similar errors, and likewise, St2 and St3 are similar)}.
The big {difference in reconstruction loss between `Scratch' and `pFEED'} in Figure \ref{rec_loss_kd} also {corresponds to} the high performance increase in ResNet-56 row of Table \ref{pfeed}. 
A better teacher network would learn more complex and detailed features because it {has} to contain powers to distinguish important {but} different details from each image to form better decision boundaries.
{It is worth noting that the curves for KD and pFEED shows opposite aspect, even though they both succeed in enhancing their performance. 
Here, our assumptions also holds: 
First, training the student with multiple teacher's feature map will help the student learn detailed features. 
Second, teacher's labels are abstract information, but will help the student learn key information.}
Thus, the tendency of KD in Figure \ref{rec_loss_kd} is opposite to pFEED.
{The curves for other type of networks also show consistent aspect, and more examples are handled in the {supplementary} materials.}

\section{Discussion}
Experiment in Section \ref{ex:parallel} with pFEED shows that allocating nonlinear transformations for each of teacher networks can extract ensemble knowledge from multiple teachers. 
{On the other hand,} AT, which directly mimics the attention map, fails to learn from multiple teacher networks. 
The error of pFEED comes closer to {the} actual model ensemble compared to KD, especially for the models with high capacity.
Next, experiments in Section \ref{sec:compare} compares sFEED with FT, KD, and BAN. Results give lessons to the choice of the algorithm that would be useful depending on the type of networks. 
The FT which extract key information from the teacher network performs better than sequential FEED for smaller networks and worse for larger networks. 
The KD and BANs, {using} labels which is even more abstract, {perform} better than sequential FEED for smaller networks. 
However, the result shows that sFEED with nonlinear transformation layers are more useful for networks {with a} larger capacity.
Though not absolute, if one wants to use distillation for model compression with smaller networks, it would be beneficial to seek ones that use abstract information like labels.
If one wants to use distillation for high performance where higher performance is needed, distillation methods that can give more detailed information can be useful.

The analysis of the reconstruction error, which we utilized the paraphraser of FT, would be helpful to judge whether the network compactly learned its features.
Finally, since we did not actively search better options for the nonlinear transformations, better choices for them might exist.

\section{Conclusion}
In this work, we proposed a couple of new network training algorithms referred to as \textit{FEature-level Ensemble {for} knowledge Distillation} (FEED). 
With FEEDs, we can improve the performance of a network by trying to inject ensemble knowledge {in the feature-map level} to the student network. 
The first one, parallel FEED trains the student network using multiple teachers simultaneously. 
The second one, sequential FEED recursively trains the student network and incrementally improves performance. 
The qualitative analysis with reconstruction loss gives hints about the cause of accuracy gains.
{The main drawback is the training times needed for multiple teachers which is an inherent characteristics of any ensemble methods,} and pFEED causes bottleneck by feeding inputs to multiple teachers simultaneously. 
But, it does not affect the inference, which is beneficial without trade-offs in {the} test time.
{Devising a more train-efficient method} will be our future work, together with an application to other domains {other than classification tasks}.

\bibliographystyle{aaai}
\bibliography{FEED_AAAI}
\end{document}